\def\year{2020}\relax
\documentclass[letterpaper]{article} 
\usepackage{aaai20}  
\usepackage{times}  
\usepackage{helvet} 
\usepackage{courier}  
\usepackage[hyphens]{url}  
\usepackage{graphicx} 
\usepackage{graphicx}  
\usepackage{amssymb}
\usepackage{mathtools}
\usepackage{multirow}
\usepackage{makecell}
\usepackage{enumitem}

\title{ECGadv: Generating Adversarial Electrocardiogram to \\ Misguide Arrhythmia Classification System}
\author{
Huangxun Chen$^1$$^\dagger$,
Chenyu Huang$^1$$^\dagger$,
Qianyi Huang$^2$$^1$,
Qian Zhang$^1$,
Wei Wang$^3$
\\ 
$^1$ The Hong Kong University of Science and Technology\\
$^2$ Southern University of Science and Technology, Peng Cheng Laboratory\\
$^3$ Huazhong University of Science and Technology \\
\{hchenay, chuangak\}@connect.ust.hk,
huangqy@sustech.edu.cn,
qianzh@cse.ust.hk,
weiwangw@hust.edu.cn \\
$^\dagger$ Co-primary Authors
}
 
\begin{document}

\maketitle

\begin{abstract}
Deep neural networks (DNNs)-powered Electrocardiogram (ECG) diagnosis systems recently achieve promising progress to take over tedious examinations by cardiologists. 
However, their vulnerability to adversarial attacks still lack comprehensive investigation. The existing attacks in image domain could not be directly applicable due to the distinct properties of ECGs in visualization and dynamic properties. Thus, this paper takes a step to thoroughly explore adversarial attacks on the DNN-powered ECG diagnosis system. We analyze the properties of ECGs to design effective attacks schemes under two attacks models respectively. 
Our results demonstrate the blind spots of DNN-powered diagnosis systems under adversarial attacks, 
which calls attention to adequate countermeasures.
\end{abstract}

\section{Introduction}

In common clinical practice, the ECG is an important tool to diagnose a wide spectrum of cardiac disorders, 
which are the leading health problem and cause of death worldwide by statistics~\cite{CVD}. 
There are recent high-profile examples of Deep Neural Networks (DNNs)-powered approaches achieving parity with human cardiologists on ECG classification and diagnosis~\cite{Hannun2019Cardiologist,aivsdoctor,kiranyaz2016real,al2016deep}, which are superior to traditional classification methods.
Given enormous costs of healthcare, it is tempting to replace expensive manual ECG examining of cardiologists with a cheap and highly accurate deep learning system. In recent, the U.S. Food and Drug Administration has granted clearance to several deep learning-based ECG diagnostic systems such as AliveCor\footnote{\url{https://www.prnewswire.com/news-releases/fda-grants-first-ever-clearances-to-detect-bradycardia-and-tachycardia-on-a-personal-ecg-device-300835949.html}} and  Biofourmis\footnote{\url{https://www.mobihealthnews.com/content/fda-clears-biofourmis-software-ecg-based-arrhythmia-detection}}.

With DNN's increasing adoption in ECG diagnosis, its potential vulnerability to `adversarial examples' also arouses great public concern. 
The state-of-the-art literature has shown that to attack a DNN-based image classifier, an adversary can construct adversarial images by adding almost imperceptible perturbations to the input image. This misleads DNNs to misclassify them into an incorrect class~\cite{szegedy2013intriguing,goodfellow6572explaining,carlini2017towards}. 
Such adversarial attacks would pose devastating threats to the DNN-powered ECG diagnosis system. On one hand, adversarial examples fool the system to give incorrect results so that the system fails to serve the purpose of diagnosis assistance. On the other hand, adversarial examples would breed medical frauds. The DNNs' outputs are expected to be utilized in other decision-making in medical system~\cite{finlayson2018adversarial}, including billing and reimbursement between hospitals/physicians and insurance companies. Large institutions or individual actors may exploit the system's blind spots on adversarial examples to inflate medical costs (e.g., exaggerate symptoms) for profit\footnote{\url{https://www.beckershospitalreview.com/legal-regulatory-issues/cardiologist-convicted-in-fountain-of-youth-billing-fraud-scam.html}}. 

To our knowledge, previous literature on DNN model attacks mainly focus on the image domain, and has yet to thoroughly discuss the adversarial attacks on ECG recordings. 
In this paper, we identify the distinct properties of ECGs, and investigate two types of adversarial attacks for DNN-based ECG classification system. 

\begin{figure}[h]
	\centering
	\includegraphics[width=0.32\textwidth]{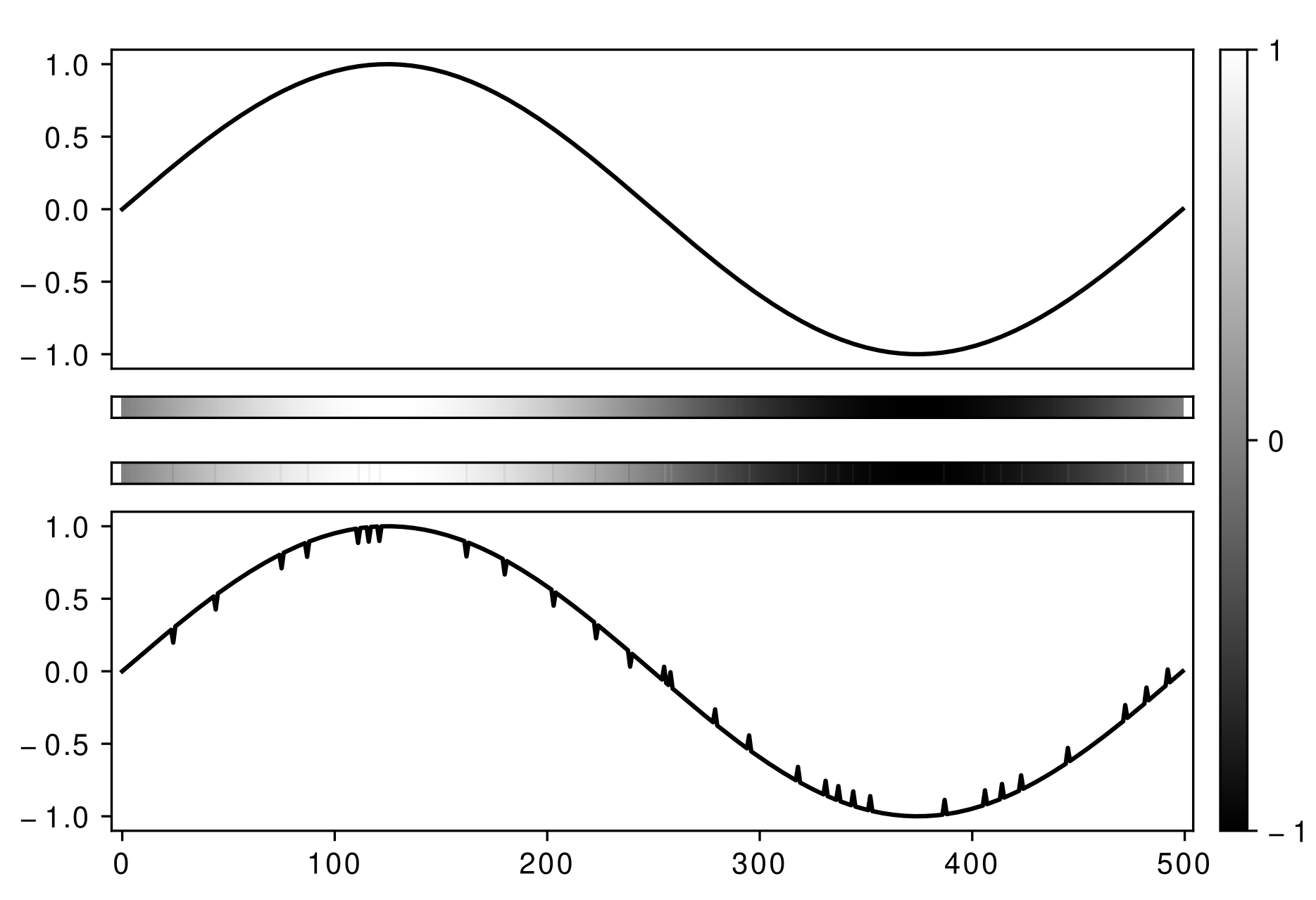}
	\caption{Perception test. There are two data arrays in the range of $[-1, 1]$, and the second one is obtained by adding a few perturbations with $0.1$ amplitude to the first one. Both of them are visualized as line curves and gray-scale images.}
	\label{fig:grayscale}
\end{figure}

In Type I Attack, the adversary can access the ECG recordings and corrupt them by adding perturbations. 
One possible case is a cardiologist who can access patients' ECGs and have monetary incentive to manipulate them to fool the checking system of insurance companies. 
Another possible case is a hacker who intercept and corrupt data to attack a cloud-deployed ECG diagnosis system for fun or profit. That data may be uploaded from portable patches like Life Signal LP1100 or household medical instruments like Heal Force ECG monitor 
to the cloud-deployed algorithms for analysis. 
For both cases, the adversary aims to engineer ECGs so that the ECG classification system is mislead to give the diagnosis that he/she desires, and in the meanwhile, the data perturbations should be sufficiently subtle that they are either imperceptible to humans, or if perceptible, seems natural and not representative of an attack.
We found that simply applying existing image-targeted attacks on ECG recordings generates suspicious adversarial instances, because commonly-used $L_p$ norm in image domain to encourage visual imperceptibility is unsuitable for ECGs (see Figure~\ref{fig:sample}). 
In visualization, each value in an ECG represents the voltage of a sampling point which is visualized as a line curve. Meanwhile, each value in a image represents the grayscale or RGB value of a pixel which is visualized as the corresponding color. Humans have different perceptual sensitivities to colors and line curves. As shown in  Fig.~\ref{fig:grayscale}, when two data arrays are visualized as line curves, their differences are more prominent rather than those visualized as gray-scale images. In this paper, we propose smoothness metrics to quantify perceptual similarities of line curves, and leverages them to generate unsuspicious adversarial ECG instances.

\begin{figure}[h]
	\centering
	\includegraphics[width=0.46\textwidth]{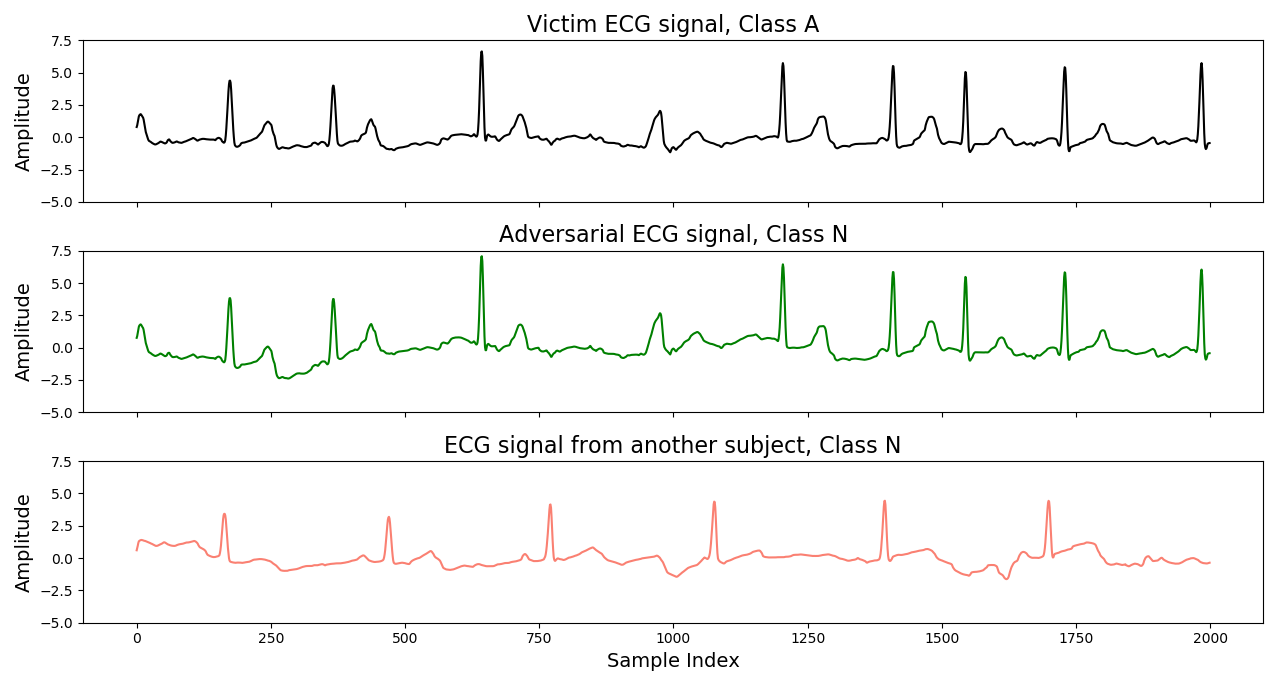}
	\caption{Adversarial Attack v.s. Substitution attack}
	\label{fig:sub}
\end{figure}

It is worth mentioning the difference between adversarial attacks and simple substitution attacks. In substitution attack, the adversary replaces the victim ECG with ECG of another subject with the target class. However, the ECGs, as a kind of biomedical signs, are often unique to their owners as fingerprint~\cite{odinaka2012ecg}. Thus, the simple substitution attacks can be effectively defended if the system checks input ECGs against prior recordings from the same patient.  However, the adversarial attacks only add subtle perturbations without substantially altering the personal identifier (Figure~\ref{fig:sub}).

In Type II Attack, the adversary may not be able to access the ECGs directly or they want to fool the system without leaving digital tampering footage. Thus, the attackers inject perturbation to on-the-fly ECGs by physical process. The feasibility of such attacks can be achieved by EMI signal injection as in~\cite{kune2013ghost}, which meant to pollute ECGs but did not consider crafting injected signals for adversarial attacks. Due to lack of equipment like patient simulator, we could not implement their prototype to conduct physical attacks. However, we identify four major properties of such attack that different from Type I, explicitly consider them in attacking strategy and mimic them in evaluation. 

\begin{enumerate}[topsep=0pt]
	\item There is skewing in time domain between perturbation and ECG, since the attacker hardly knows ECG’s exact start time. We mimic it by shifting perturbation with various amount before adding to victim ECGs.
	\item Filtering of ECG devices, as a standard process to combat noise, will be applied on injected perturbation and may impair its attack effect. In evaluation, two widely-adopted filters are applied to adversarial examples. The rectangular filter is used in generation process, since it strictly removes all power within selected frequency range.
	\item Type I accesses digital ECGs without showing up on the scene, but physical injection should be conducted closer to victim. The smaller attack duration(the part of an ECG affected by perturbation), the lower exposure risk. Thus we generate perturbation with different attack duration for evaluation.
	\item Physical attacks inject perturbation generated from known ECGs to on-the-fly ECGs. The variance between them may affect attack effect. In evaluation, perturbation generated with random selected ECGs are tested on others.
\end{enumerate}

In summary, the contributions of this paper are as follows:
\begin{itemize}
	\item This paper thoroughly investigate adversarial attacks for DNN-based ECG classification systems. We identify the distinct properties of ECGs to facilitate designing effective attack schemes under two attack models respectively.  
	\item We propose a smoothness metric to effectively quantify human perceptual distance on line cures, which quantifies the pattern similarity in a computationally-efficient way. Adversarial attacks using the smoothness metric achieve a 99.9\% success attack rate. In addition, we conduct an extensive human perceptual study on both ordinary people and cardiologists to evaluate the imperceptibility of adversarial ECG instances.
	\item We model the sampling point uncertainty of the on-the-fly ECGs and the filtering effect within the adversarial generation scheme. The generated perturbations are skewing-resistant and filtering-resistant to tamper with on-the-fly signals (99.64\% success rate), and generalize well in unseen examples.
\end{itemize}

\section{Related Works}

Here we review recent works on adversarial examples, and the existing arrhythmia classification systems.

\subsection{Adversarial Examples}
Recently, considerable attack strategies have been proposed to generate adversarial examples. 
Attacks can be classified into targeted and untargeted ones based on the adversarial goal. The adversary of the former modifies an input to mislead the targeted model to classify the perturbed input into a chosen class, while the adversary of the latter make the perturbed input misclassified to any class other than the ground truth. In this paper, we only focus on the more powerful targeted attacks. 

Based on the accessibility to the target model, the existing attacks fall into white-box and black-box attacks categories. In former manner, an adversary has complete access to a classifier~\cite{szegedy2013intriguing,goodfellow6572explaining,moosavi2016deepfool,carlini2017towards,kurakin2018adversarial}, while in latter manner, an adversary has zero knowledge about them~\cite{papernot2016transferability,moosavi2017universal,liu2016delving}. This paper studies the white-box adversarial attacks to explore the upper bound of an adversary to better motivate defense methods. Besides, prior works~\cite{papernot2016transferability,liu2016delving} have shown the transferability of adversarial attacks, i.e, train a substitute model given black-box access to a target model, and transfer the attacks to it by attacking the substitute one.

Adversarial attacks have been studied most in image domain. However, in other domains, these attack schemes may lose effect, e.g., \cite{qin2019imperceptible} identifies unique problems on speech recognition and leverages properties of human auditory system to generate audio adversarial examples. This paper, however, focuses on adversarial attacks on ECG diagnosis, another important application domain of DNN. 

In the image domain, most works adopted $L_p$ norm as approximations of human perceptual distance to constrain the distortion. However, for ECGs in time-series format, people focus more on the overall pattern/shape, which can not be fully described by $L_p$ norm~\cite{eichmann2015evaluating,gogolou2018comparing} (see Section `Similarity Metrics' for details). 
Recent works~\cite{kurakin2018adversarial,athalye2018synthesizing,chen2018robust} have explored the robustness of the adversarial examples in the physical world, where the input images could not be precisely controlled, and may change under different viewpoints, lighting and camera noise. 
Our strategy on Type II attack is inspired by~\cite{athalye2018synthesizing,brown2017adversarial}. Different from images, we deal with sampling point uncertainty of the periodic ECGs and the filtering function of ECG devices. 

Recent works on GAN-based attacks~\cite{xiao2018generating,song2018constructing} focus on improve attacking efficiency to image classification system, which can be combined with metric computation efficiency of ECGadv in future work. 
A workshop paper~\cite{hanadversarial} convolves perturbation with Gaussian kernels for ECG adversarial attacks. Our proposed smoothness metric and Gaussian kernels method can be integrated to improve the system. Besides, our paper further addresses the issues in physical ECG attacks. 
For the emerging defense methods, ~\cite{athalye2018obfuscated} proposed a general framework to circumvent several published defenses based on randomly transforming the input. Thus, we do not discuss defense breaking in this paper.

\subsection{Arrhythmia Classification System}
Considerable efforts have been made on automated arrhythmia classification systems to take over tedious manual examinations.
Deep learning methods show great potential
due to their ability to automatically learn features through multiple levels of abstraction, which frees the system from the dependence on hand-engineered features. 
Recent works~\cite{kiranyaz2016real,al2016deep,Hannun2019Cardiologist}  started applying DNN models on ECG signals for arrhythmia classification and achieved good performance.
For any system in the health-care field, it is crucial to defend against any possible attacks since people's lives rely heavily on the system's reliability. Prior work~\cite{kune2013ghost} has launched attacks to pollute the measurement of cardiac devices by a low-power emission of chosen electromagnetic waveforms. 
The adversarial attacks and the injection attacks in~\cite{kune2013ghost} complement each other. The injection attack can inject the carefully-crafted perturbation generated by adversarial attacks to perform targeted attacks to mislead the arrhythmia classification system.

\section{Technical Approach}
In this section, we illustrate our attack strategies for two threat models respectively.
\subsection{Type I Attack Strategy}
\label{sec:metric}
\subsubsection{Problem Formulation}
Given an m-class classifier, $g : \mathcal{X} \to \mathcal{Y}$ that accepts an input $x  \in \mathcal{X}$ and produces an output $y \in \mathcal{Y}$.  The output vector $y$, treated as the probability distribution, satisfies $0 \leq y_i \leq 1$ and $\sum_{i=1}^{m} y_i = 1$. The classifier assigns the label $\mathit{C}(x) = \mathrm{argmax}_i  g(x)_i$ to the input $x$. 
Let $\mathit{C}^*(x)$ be the correct label of $x$. 
Given a valid input $x$ and a target class $t \ne \mathit{C}^*(x)$, an adversary aims to generate adversarial examples $x_{adv}$ so that the classifier predicts $g(x_{adv}) = t$ (\textit{i.e.} successful attack),  and $x_{adv}$ and $x$ are close based on the similarity metric (\textit{i.e.} visual imperceptibility). 
It can be modeled as a constrained minimization problem as seen in prior works~\cite{szegedy2013intriguing}:
\begin{equation}
\begin{split}
& \mathrm{minimize} \text{  } \mathcal{D}(x, x_{adv}) \\
& \mathrm{such} \text{ } \mathrm{that} \text{  }  \mathit{C}(x_{adv}) = t 
\end{split}
\end{equation}
where $\mathcal{D}$ is some similarity metric. It is worth mentioning that there is no box constraints for time-series measurement.
It is equivalent to solve~\cite{carlini2017towards}:
\begin{equation}
\begin{split}
& \mathrm{minimize} \text{  } \mathcal{D}(x, x_{adv}) + c \cdot \mathit{f}_{g}(x_{adv}) \\
\end{split}
\end{equation}
where $ \mathit{f}_{g}$ is an objective function mapping the input to a positive number, which satisfies $\mathit{f}_{g}(x_{adv}) \leq 0$ if and only if $ \mathit{C}(x_{adv}) = t$.
One common objective function is cross-entropy. We adopt the one in~\cite{carlini2017towards}.
\begin{equation}
\mathit{f}_{g}(x_{adv}) = (\mathrm{max}_{i \ne t}(Z(x_{adv})_i) - Z(x_{adv})_t)^+ 
\end{equation}
where $Z(x) = z$ is logits, \textit{i.e.}, the output of all layers except the softmax. $(e)^+$ is short-hand for $\mathrm{max}(e, 0)$.

\subsubsection{Similarity Metrics}
\label{sec:metric}
To generate adversarial examples, we require a distance metric to quantify perceptual similarity to encourage visual imperceptibility. The widely-adopted distance metrics in the literature are $L_p$ norms $\Vert  x_{adv}-x \Vert _p$, where the p-norm  $\Vert  \cdot \Vert_p$ is defined as 
$\Vert  v \Vert _p = (\sum^n_{i=1} \vert v_i \vert ^p) ^{\frac{1}{p}}$. 
$L_p$ norms focus on the change in each pixel value. However, human perception on line curves focuses more on the overall pattern/shape. Studies in ~\cite{eichmann2015evaluating,gogolou2018comparing} show that given a group of line curves for similarity assessment, pattern-focused distance metrics like the Dynamic time warping (DTW)-based ones produce rankings that are closer to the human-annotated rankings than value-focused metrics like Euclidean distances. 
Thus, we consider using DTW to quantify the similarity of ECGs at first. However, the non-differentiability and non-parallelism of DTW make it ill-suited for adversarial attacks. 
Recent work~\cite{cuturi2017soft} proposes a differentiable DTW variant, Soft-DTW. However, Soft-DTW does not change the essence of DTW -- a standard dynamic programming problem. The value and gradient of Soft-DTW would be computed in quadratic time, and it is hard to leverage the parallel computing of the GPU to speed it up. 
To capture the pattern similarity in a computation-efficient way, we adopt the following metric, denoted as \textit{smoothness} as our similarity metric. Given $\delta = x_{adv} - x $ and $var(\cdot)$ refers to variance calculation:

\begin{equation}
\label{eq:smooth}
\begin{split}
& \mathit{diff}(\delta) = {\delta_i - \delta_{i-1}}, i=2, \dots ,n \\
& d_{\mathrm{smooth}}(\delta) = \mathrm{var}(\mathit{diff}(\delta)) \\
\end{split} 
\end{equation}

Smoothness metric $d_\mathrm{smooth}$ quantifies the smoothness of perturbation($\delta$) by measuring the variation of the difference between neighbouring points of perturbation. 
The smaller the variation, the smoother the perturbation. A smoother perturbation $\delta$ means that the adversarial instances $x'$ are more likely to preserve a similar pattern to the original instance $x$. In the extreme case where $d_\mathrm{smooth} = 0$, $\delta$ should be a constant and $x_{adv} = x + \mathrm{constant}$, \textit{i.e.}, the adversarial instances $x_{adv}$ have the same shape as the original instance $x$.
It is worth mentioning that in our attack scheme, we intentionally preserve the zero-mean and one-variance property of the generated $x_{adv}$, therefore the perturbation can not be easily filtered by the normalization layer of the system. 
Besides, compared with the quadratic time complexity of Soft-DTW, the smoothness metric can be computed in linear time, which is efficient in principle. To further quantify the efficiency, we run the adversarial attacks with different metrics: Soft-DTW, smoothness metric and $L_2$ norm. Both the computing resources (AWS c5.2xlarge instances) and the victim ECGs are the same. The average CPU time per iteration of different metrics are shown in Table~\ref{tab:metriccomp}. The smoothness metric can be further accelerated by GPU.

\renewcommand{\arraystretch}{1.1}
\begin{table}[h]
	\begin{center}
		\begin{tabular}{ |c|c|c|c|} 
			\hline
			Metric & $d_{\mathrm{softdtw}}$ & $d_{\mathrm{smooth}}$& $d_{\mathrm{l2}}$ \\
			\hline
			CPU time/iteration & 12.28s & 0.05s & 0.05s \\
			\hline
		\end{tabular}
	\end{center}
	\caption{Computation Efficiency across Different Metrics}
	\label{tab:metriccomp}
\end{table}

\subsection{Type II Attack Strategy}
\label{sec:local}
\subsubsection{Problem Formulation}
Given the same m-class classifier, $g : \mathcal{X} \to \mathcal{Y}$ as above, in Type II attack, 
we explicitly consider the filtering process in attack scheme. Filtering is a standard process in ECG devices to combat noises before the data analysis, including baseline wandering noises ($<$0.05Hz) and the power-line noises (50 or 60 Hz)~\cite{luo2010review}. To generate filtering-resistant perturbations, we constrain the power of the perturbation within those filtered frequency bands during the optimization procedure.
We also consider the possible skewing to generate perturbations that are effective for the on-the-fly ECGs, since it is hard for the attacker to obtain the exact time that the device begins measuring ECGs.
Inspired by \textit{Expectation Over Transformation(EOT)}~\cite{athalye2018synthesizing}, we regard such uncertainty as a shifting transformation of the original measurement and explicitly consider such a transformation within the optimization procedure.

Formally, given a distance function $\mathcal{D(\cdot, \cdot)}$ and a chosen distribution $\mathit{T}$ of transformation function $t$, we have the following optimization problem:
\begin{equation}
\begin{split}	
\mathrm{minimize} \text{ } &\mathbb{E}_{t \sim T}[\mathcal{D}(t(x_{adv}), t(x))] + \\
& c \cdot \mathbb{E}_{t \sim T}[-\log P(y_t|t(x_{adv}))]\\
\end{split}
\end{equation}
where $x_{adv} = x + h(x_{perturb})$. $x_{perturb}$ is the added perturbation and $h(\cdot)$ is a rectangular filter. Specifically, we transform the $x_{perturb}$ from time domain to frequency domain via Fast Fourier transform. We utilize a mask to zero the power of frequency bins for less than 0.05Hz and 50/60Hz. Finally, inverse Fast Fourier transform will transform it back to the time domain.
Besides, we add a constraint $ \epsilon_1 < \mathbb{E}_{t \sim T}[\mathcal{D}(t(x_{adv}), t(x))] < \epsilon_2$. $\epsilon_1$ is large enough that $x_{adv}$ can have a large probability of successful attacks under most shifting transformations. Since the ECG signals of the same class share common pattern, a sufficiently large $\epsilon_1$ can implicitly enable the universality of an adversarial sample, \textit{i.e.}, a perturbation is effective on other unseen samples of the same class. $\epsilon_2$ forces the adversarial examples to be within a certain distance constraint of the original. 

\subsubsection{Perturbation Window Size}
For adversarial attacks, it is better that the perturbation attracts minimal attention of the victim. Thus, we introduce the length of the perturbation $w_d$ as a parameter, which could be set by the adversary and fixed during the perturbation generation. 
$w_d$ gives the system flexibility to control the added perturbation. The intuition behind is that the smaller $w_d$ is, the smaller the attack duration. Attack duration denotes the time when the attacker try to inject the signal. It is obviously that the less time the attacker stays active in the crime scene, the less chance it will be perceived by the victim.
Moreover,  the larger $w_d$ is, the generated perturbation has higher probability of having an effect on other unseen samples of the same class(\textit{i.e.}, universality).

\section{Experimental Results}

In this section, we first introduce the victim DNN-based ECG classification system for attack scheme evaluation, then evaluate our attacks in two threat models respectively\footnote{\url{https://github.com/codespace123/ECGadv}}.

\subsection{Victim DNN-powered ECG Diagnosis Model}
\label{sec:targetmodel}
We apply our attack strategies to the DNN-based arrhythmia classification system~\cite{rajpurkar2017cardiologist,andreotti2017comparing,Hannun2019Cardiologist}. 
An arrhythmia is defined as any rhythm other than a normal rhythm. 
If the detection algorithm is mislead to classify an arrhythmia as a normal one, the patient may miss the optimal treatment period. Conversely if a normal rhythm is misclassified as an arrhythmia, the patient may accept unnecessary consultation and treatment, which results in medical resources waste or frauds. 

The original model~\cite{rajpurkar2017cardiologist} adopts 34-layer Residual Networks (ResNet)~\cite{he2016deep} to classify a 30s single-lead ECG segment into 14 different classes. However, their dataset and trained model are not public. In the Physionet/Computing in the Cardiology Challenge 2017~\cite{clifford2017af}, \cite{andreotti2017comparing} reproduced the approach by ~\cite{rajpurkar2017cardiologist} on the PhyDB dataset and achieved a good performance. The model is the representative of the current state-of-the-art in arrhythmia classification. Both their algorithm and model are available in open-source\footnote{\url{https://github.com/fernandoandreotti/cinc-challenge2017}}. 
PhyDB dataset consists of 8,528 short single-lead ECG segments labeled as 4 classes: normal rhythm(N), atrial fibrillation(A), other rhythm(O) and noise($\backsim$). Both atrial fibrillation and other rhythm indicates arrhythmia. Atrial fibrillation is the most prevalent cardiac arrhythmia. ``Other rhythm'' in the dataset refers to other abnormal arrhythmia except atrial fibrillation.
For note, the accuracy of this model is not 100\% on the PhyDB dataset. Thus, to prove the effectiveness of the proposed attacks, we only generate adversarial examples for those ECGs originally correctly classified by the model without attacks. The profile of the attack dataset is shown in Table~\ref{tab:profile} (6081 ECGs in total). The sampling rate of the ECGs is 300Hz, \textit{i.e.}, the length of a 30s ECG is 9000.

\subsection{Evaluation for Type I Attack}
\label{sec:cloud_eval} 
\subsubsection{Experiment Setup}
We implement our attack strategy for Type I Attack under the framework of CleverHans~\cite{papernot2018cleverhans}.  We adopt the Adam optimizer~\cite{kingma2014adam} with $0.005$ learning rate to search for adversarial examples. 
We compare the performance of three similarity metrics on adversarial examples generation, given $\delta = x_{adv} - x$: (i) $d_\mathrm{l2}(\delta) = \Vert \delta \Vert_2^2$, (ii)$d_{\mathrm{smooth}}(\delta)$ (Equation~\ref{eq:smooth}), (iii) $d_\mathrm{smooth, l2}(\delta)$ = $d_\mathrm{smooth}(\delta) + k \cdot d_\mathrm{l2}(\delta)$, $k=0.01$.
All metrics are evaluated under the same optimization scheme with the same hyper-parameters. 

\begin{table}[h]
	\begin{center}
		\begin{tabular}{ |c|c|c|c|} 
			\hline
			\multirow{2}{*}{Type} & \multirow{2}{*}{Number} & \multicolumn{2}{|c|}{Time length (s)}  \\
			\cline{3-4}
			& & mean & std \\
			\hline
			Normal rhythm(N)& 3886 & 32.85 & 9.70 \\ 
			Atrial Fibrillation(A) & 447 & 32.25 & 11.98\\
			Other rhythm(O) & 1488 & 35.46 & 11.56\\
			Noisy signal($\backsim$) & 260 & 24.02  & 10.42 \\
			\hline
		\end{tabular}
	\end{center}
	\caption{Data profile for the attack dataset}
	\label{tab:profile}
\end{table}

\renewcommand{\arraystretch}{1.2}
\begin{table*}[h]
	\begin{center}
		\begin{tabular}{ |c|c|c|c|c|c|c|c|c|c|c|c|c|c|} 
			\hline
			\multirow{2}{*}{}  & \multicolumn{4}{|c|}{$d_\mathrm{l2}$} & \multicolumn{4}{|c|}{$d_\mathrm{smooth}$} & \multicolumn{4}{|c|}{$d_\mathrm{smooth, l2}$}  \\
			\cline{2-13}
			& A & N & O & $\backsim$ & A & N & O & $\backsim$ & A & N & O & $\backsim$  \\
			\hline
			A & / & 97.22\% & 100\% & 100\% & / & 100\% & 100\% & 100\% & / & 100.0\% & 100.0\% & 100.0\% \\
			\hline
			N & 100\% & / & 100\% & 100\% & 100\% & /& 100\% & 100\% & 100\% & / & 100\% & 100\% \\
			\hline
			O & 99.44\% & 95.0\% & / & 100\% & 99.72\% & 100\% & / & 100\% & 100\% & 100\% & / & 100\% \\
			\hline
			$\backsim$ & 100\% & 99.55\% & 100\% & / & 100\% & 100\% & 100\% & / & 100\% & 100\% & 100\% & / \\
			\hline
		\end{tabular}
	\end{center}
	\caption{Success rates of targeted attacks (Type I Attack)}
	\label{tab:success}
\end{table*}

\subsubsection{Success Rate of Targeted Attacks}
We select the first 360 segments of class N, class A and class O respectively, and the first 220 segments of class $\backsim$ in attack dataset to evaluate the success rate of the targeted attacks. For each ECG segment, we conduct three targeted attacks to other classes one by one. Thus, we have 12 source-target pairs given 4 classes. The attack results are shown in Table~\ref{tab:success}. 
With all three similarity metrics, the generated adversarial instances achieve high attack success rates. $d_\mathrm{l2}$ fails in a few instances of some source-target pairs, such as ``O $\rightarrow$ A'', ``A $\rightarrow$ N'', ``O $\rightarrow$ N'' and ``$\backsim$ $\rightarrow$ N''. $d_{\mathrm{smooth}}$ case achieves almost a 100\% success rate and $d_\mathrm{smooth, l2}$ achieves a 100\% success rate. 

\begin{figure*}[h]
	\begin{minipage}{0.61\textwidth}
		\centering
		\includegraphics[width=0.98\textwidth]{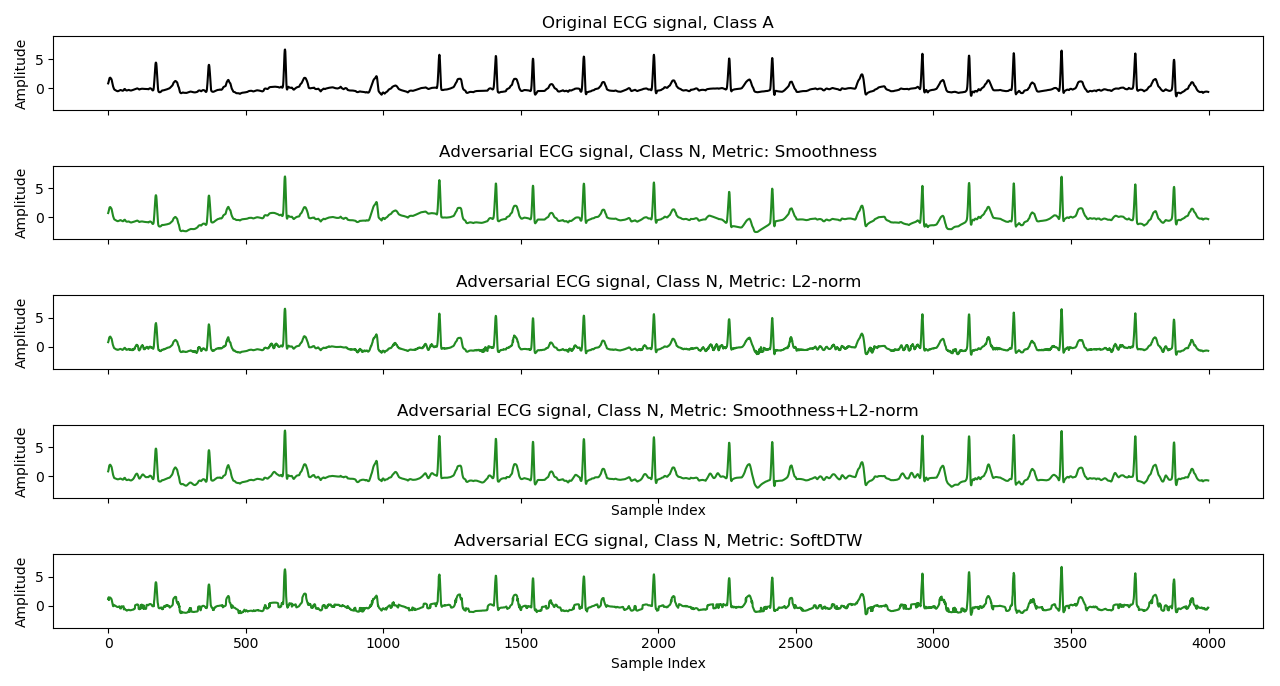}
		\caption{A sample of generated adversarial ECG signal.}
		\label{fig:sample}
	\end{minipage}
	\begin{minipage}{0.39\textwidth}
		\centering
		\includegraphics[width=\textwidth]{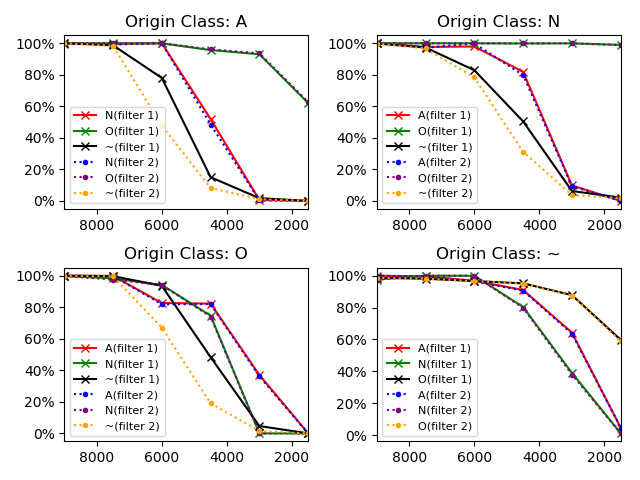}
		\caption{Success attack rates with different sized windows.}
		\label{fig:wd}
	\end{minipage}
\end{figure*}

A sample of generated adversarial ECG signals are shown in Fig.~\ref{fig:sample}. Due to the limited space, we only show a case where an original atrial fibrillation ECG(A) is misclassified to a normal rhythm(N). Compared with original ECG, $d_\mathrm{l2}$ one presents small but consecutive peaks at multiple locations, which are almost impossible in cardiac rhythms. While the $d_\mathrm{smooth}$ one presents smooth signal transition and preserves more similar pattern to the original. 
It is also noticed that the Soft-DTW one present suspicious spikes, the extent of which falls in between L2-norm and `smoothness+L2' cases. 
Table~\ref{tab:metriccomp} shows CPU time per iteration of softDTW is about 12 seconds, and it takes hundreds of iterations to generate one example. It is time-consuming to generate large number of them (600 in our perceptual study) using softDTW. Since the proposed attacks as shown in Table~\ref{tab:success} almost achieve 100\% success rate, without affecting major conclusions, we exclude softDTW in evaluation.

\subsubsection{Human Perceptual Study}
\label{sec:amt}
We conduct an extensive human perceptual study on both ordinary people and cardiologists to evaluate the imperceptibility of adversarial ECGs.

Ordinary human participants without medical expertise are recruited from Amazon Mechanical Turk(AMT).  Thus, they are only required to compare the adversarial examples generated using different similarity metrics and choose the one closer to the original ECG. 
For each similarity metric, we generate 600 adversarial examples (each source-target pair accounts for 50 examples). In the study, the participants are asked to observe an original example and its two adversarial ones generated using two different similarity metrics. Then they need to choose one of the two adversarial examples that is closer to the original.
The perceptual study comprises three parts, (\romannumeral 1) $d_\mathrm{smooth}$ versus $d_\mathrm{l2}$, (\romannumeral 2) $d_\mathrm{smooth, l2} $ versus $d_\mathrm{l2}$, and (\romannumeral 3) $d_\mathrm{smooth}$ versus $d_\mathrm{smooth, l2}$. 
To avoid labeling bias, we allow each user to conduct at most 60 trials for each part. For each tuple of an original example and its two adversarial examples, we collect 5 annotations from different participants. 
In total, we collected 9000 annotations from 57 AMT users. The study results are shown in Table \ref{tab:subjective}, where ``triumphs" denotes the metric got 4 or 5 votes for all 5 annotations, and ``wins'' denotes that the metric got 3 votes for 5, \textit{i.e.}, a narrow victory.

\renewcommand{\arraystretch}{1.2}
\begin{table}[h]
	\begin{center}
		\resizebox{\linewidth}{!}{%
		\begin{tabular}{ |c|c|c|c|c|c|c|} 
			\hline
			\multirow{3}{*}{\romannumeral 1} & \multicolumn{3}{|c|}{$d_\mathrm{smooth}$ wins(\%)} & \multicolumn{3}{|c|}{ $d_\mathrm{l2}$ wins(\%)} \\ 
			\cline{2-7}
			& triumphs & wins &  total & triumphs & wins &  total \\
			\cline{2-7}
			& 58.67 & 22.67 & 81.34 & 10 & 8.66 &  18.66 \\
			\hline
			\hline
			\multirow{3}{*}{\romannumeral 2} & \multicolumn{3}{|c|}{$d_\mathrm{smooth, l2}$ wins(\%)} & \multicolumn{3}{|c|}{ $d_\mathrm{l2}$ wins(\%)} \\ 
			\cline{2-7}
			& triumphs & wins &  total & triumphs & wins &  total \\
			\cline{2-7}
			& 65.5 & 18.5 & 84 & 7.83 & 8.17 &  16 \\
			\hline
			\hline
			\multirow{3}{*}{\romannumeral 3} & \multicolumn{3}{|c|}{$d_\mathrm{smooth}$ wins(\%)} & \multicolumn{3}{|c|}{ $d_\mathrm{smooth, l2}$ wins(\%)} \\ 
			\cline{2-7}
			& triumphs & wins &  total & triumphs & wins &  total \\
			\cline{2-7}
			& 31.83 & 27.83 & 59.67 & 15.83 & 24.5 &  40.33\\
			\hline
		\end{tabular}}
	\end{center}
	\caption{Human perceptual study (AMT participants)}
	\label{tab:subjective}
\end{table}

Compared with the $d_\mathrm{l2}$-generated examples, the $d_\mathrm{smooth}$-generated ones are voted closer to the original in 81.34\% of the trials. 
This indicates that the smoothness metric encourages generated adversarial examples preserve similar patterns to original ones, so they are more likely to be imperceptible. 
When comparing $d_\mathrm{smooth}$ and $d_\mathrm{smooth, l2}$, $d_\mathrm{smooth}$ get a few more votes (59.67\%) than $d_\mathrm{smooth, l2}$, which further validates that the smoothness metric better qualifies human similarity perception on line curves than $L_2$ norm. 
The data provide sufficient evidence (p values $<$ 0.0001 using z-test) at the 5\% level of significance to conclude that most people think $d_\mathrm{smooth}$ is more imperceptible than $d_\mathrm{l2}$ and $d_\mathrm{smooth, l2}$.

Besides participants on AMT, we also invite three cardiologists to evaluate whether added perturbations arouse their suspicion.
The cardiologists are asked to classify the given ECG and its adversarial counterparts into 4 classes(A, N, O, $\backsim$) based on their medical expertise. 
We focus on the cases of ``N $\rightarrow$ A'', ``N $\rightarrow$ O'', ``A $\rightarrow$ N'', ``O $\rightarrow$ N'', which misclassify a normal rhythm to an arrhythmia or vise versa.  For the above 4 source-target pairs, we randomly select 6 type N, 3 type A and 3 type O, then we generate adversarial examples with different similarity metrics. Thus, we have 48 samples (original and adversarial ones) and shuffle them randomly. For every sample, we collect annotations from all three cardiologists. The results are shown in Table~\ref{tab:subjective_2}. 

\begin{table}[h]
	\begin{center}
		\begin{tabular}{|c|c|c|c|c|} 
			\hline
			Idx & Original & $d_\mathrm{l2}$ & $d_\mathrm{smooth}$   & $d_\mathrm{smooth, l2}$  \\
			\hline
			1 & 100\% & 100\% & 100\% & 100\% \\
			\hline
			2 & 91.7\% & 100\% & 100\% & 100\% \\
			\hline
			3 & 100\% & 100\% & 100\% & 100\% \\
			\hline
		\end{tabular}
	\end{center}
	\caption{Human Perceptual Study (Cardiologists)}
	\label{tab:subjective_2}
\end{table}

Each row refers to one cardiologist. The first column denotes the percentage of the cardiologist's annotations the same as the labels in PhyDB dataset. Only one cardiologist annotates a type A instance as type O. The last three columns show the percentage of adversarial examples which are annotated the same type as their original counterparts. 
The results show that in all cases, cardiologists give the same annotations to adversarial examples as their original counterparts. The possible reason is that most perturbations generally occur on the wave valley, but the cardiologists give annotations based on the peak-to-peak intervals. They think the subtle perturbations possibly caused by instrument noise. The results that adversarial signals can be correctly classified by cardiologists but wrongly classified by the classifier prove that our attacks successfully fool the classifier to disable its function of diagnosis assistance without arousing suspicion. 

\subsection{Evaluation for Type II Attack}
\label{sec:local_eval}
\subsubsection{Success Rate of Targeted Attacks}

We implement our attack strategy for Type II attack under the framework of CleverHans~\cite{papernot2018cleverhans}.  
During training, we maximize the objective function using the Adam~\cite{kingma2014adam} optimizer, and approximate the gradient of the expected value through independently sampling transformations at each gradient decent step. 
Among 12 source-target pairs, we randomly choose 10 samples of each pair to generate adversarial 
perturbations by applying the attack strategy in Section~\ref{sec:local}. We generate one perturbation from one sample. To generate filtering-resistant perturbation, we use rectangular filter that removes the signal with frequency of lower than 0.05Hz and 50/60Hz. Because the rectangular filter can remove all the energy within the chosen frequency band which is stricter than other filters. In this evaluation, we generate the perturbation at full length, \textit{i.e.}, $w_d$ is equal to 9000.

During testing, we apply the generated perturbations to 100 randomly-chosen samples from every source class to see whether the adversarial examples could mislead the classifier universally. By source class, we mean the chosen testing sample has the same class with training sample generating perturbation.
Before adding perturbations to the target sample, we apply a filter on the perturbations to test the filtering-resistance. The filter has two choices: Filter 1 is the rectangular filter which is the same as the training procedure. Filter 2 is the combination of two common filters used in ECG signal processing, a high-pass butterworth filter with 0.05Hz cutting frequency and notch filters for 50/60Hz power line noises. 
To mimic the sampling point uncertainty of the on-the-fly signals, we randomly shift perturbations and add them to the original signals for 200 times.The average success rates are shown in Table~\ref{tab:success_local}. The row refers to origin class and the column refers to target class. In one cell, the top success attack rate is for filter 1 and the bottom is for filter 2.
Our attack strategy achieves pretty high success rates, which indicates that the generated perturbation is filtering-resistant, skewing-resistant and universal. 

\begin{table}[h]
	\begin{center}
		\begin{tabular}{|c|c|c|c|c|} 
			\hline
			& A & N & O & $\backsim$ \\
			\hline
			A & / & \makecell{99.97\%/\\99.96\%} &  \makecell{99.82\%/\\99.86\%} &  \makecell{100\%/\\100\%} \\
			\hline
			N &  \makecell{100\%/\\100\%} & / &  \makecell{100\%/\\100\%} &  \makecell{99.83\%/\\99.75\%} \\
			\hline
			O &  \makecell{100\%/\\100\%} &  \makecell{99.76\%/\\99.73\%} & / &  \makecell{100\%/\\100\%} \\
			\hline
			$\backsim$ &  \makecell{100\%/\\100\%} &  \makecell{97.63\%/\\97.45\%} &  \makecell{98.70\%/\\98.76\%} & / \\
			\hline
		\end{tabular}
	\end{center}
	\caption{Success rates of targeted attacks (Type II Attack)}
	\label{tab:success_local}
\end{table}

\subsubsection{Impact of Window Size}

\label{sec:local_factor}
In this section, we evaluate the success attack rates with different sized windows $w_d$.
As mentioned before, the smaller the window size, the lower the chance that the attacker can be perceived. In this evaluation, we generate perturbations on different sized windows 9000,7500,6000,4500,3000 and 1500. For each window size, we generate adversarial examples under the same conditions as the previous section – randomly 10 samples for each source-target pairs. Then we apply filters, shift the perturbation randomly and add it to other samples from the original source class. The results are shown in Figure~\ref{fig:wd}. The legend refers to target class under different filters. In most cases, the success rate decreases a lot when the window size decreases. However, they slowly decrease and even remain almost unchanged under the cases of ``A $\rightarrow$ O'', ``N $\rightarrow$ O''
and ``$\backsim$ $\rightarrow$ O''. All these cases are from a certain class to class O. This is mainly because class O (refers to other abnormal arrhythmia except atrial fibrillation) may cover an expansive input space so that it is easier to misclassify an other class to class O. 
Besides, we find that except for class O, the success rate decrease more slowly when the target class is A. 
The possible reason is the inherent property of class A, \textit{i.e.}, if a certain part of the ECG signal is regraded as atrial fibrillation, then the whole ECG segment will be classified as class A. The success attack rates under different filters are quite similar, which shows the filtering-resistance of our generated perturbations. 

\section{Conclusion}
This paper proposes ECGadv to generate adversarial ECG examples to misguide arrhythmia classification systems.
The existing attacks in image domain could not be directly applicable due to the distinct properties of ECGs in visualization and dynamic properties.
We analyze the properties of ECGs to design effective attacks schemes under two attacks models respectively.
Our results demonstrate the blind spots of DNN-powered diagnosis systems under adversarial attacks to call attention to adequate countermeasures.

\subsubsection{Acknowledgment}
This work was supported in part by the RGC under Contract CERG 16203719, 16204418 and in part by the Guangdong Natural Science Foundation No. 2017A030312008.

\bibliographystyle{aaai}
\bibliography{mybib}
\end{document}